\documentclass{article}
\usepackage{spconf,amsmath,graphicx}
\usepackage{bm}
\usepackage{epsf}
\usepackage{times}
\usepackage{float}
\usepackage{graphicx}
\usepackage{tikz}
\usepackage{amsmath,amssymb,amsthm}
\usepackage{cite}
\usepackage{subcaption}
\usepackage{algorithm}
\usepackage{algorithmic}

\usepackage{float} 
\allowdisplaybreaks

\makeatletter
\newcommand{\rom}[1]{\expandafter\@slowromancap\romannumeral #1@}
\makeatother
\def\bs{\bm s}

\def\by{\bm y}

\def \bB{\bm B}

\renewcommand{\v}[1]{\ensuremath{\boldsymbol{#1}}}
\newcommand{\N}{{\ensuremath{\mathcal{N} }}}

\title{Privacy-Preserving Distributed Expectation Maximization for Gaussian Mixture model using subspace perturbation}
\name{Qiongxiu Li$^{\star}$, Jaron S. Gundersen$^{\dagger}$, Katrine Tjell$^{\dagger}$
Rafal Wisniewski$^{\dagger}$, Mads G. Christensen$^{\star}$
}
\address{
$^{\star}$ Audio Analysis Lab, CREATE, Aalborg University, Denmark, \{qili, mgc\}@create.aau.dk \\
			$^{\dagger}$ Department of Electronic Systems, Aalborg University, Denmark, \{jaron, kst, raf\}@es.aau.dk 
			}
\begin{document}
\ninept
\maketitle{}
\raggedbottom
\begin{sloppy}
\addtolength{\abovedisplayskip}{-1.0mm}
\addtolength{\belowdisplayskip}{-1.0mm}

\begin{abstract}
Privacy has become a major concern in machine learning. In fact, the federated learning is motivated by the privacy concern as it does not allow to transmit the private data but only intermediate updates. However, federated learning does not always guarantee privacy-preservation as the intermediate updates may also reveal sensitive information. In this paper, we give an explicit information-theoretical analysis of a federated  expectation maximization algorithm for Gaussian mixture model and prove that the intermediate updates can cause severe privacy leakage. To address the privacy issue, we propose a fully decentralized privacy-preserving solution, which is able to securely compute the updates in each maximization step. Additionally, we consider two different types of security attacks: the  honest-but-curious and eavesdropping adversary models. 
Numerical validation shows that the proposed approach has superior performance compared to the existing approach in terms of both the accuracy and privacy level.  
\end{abstract}
\section{Introduction}
Distributed algorithms are widely used because of their advantageous  distribution of computational burden, flexibility, and scalability of the system, and resilience because of the elimination of the potential single point of failure. Moreover, the increasing focus on individual's right to data privacy leads to even more motivation for the distributed setup. An evident example is when data is collected by multiple nodes/parties/agents and computations involve the combined dataset (to yield more accurate results). Consider for instance hospitals across cities and countries collecting data on diseases. Because of strict data protection laws, it is non-trivial for hospitals to share data with other hospitals especially if data is to travel across borders. However, performing statistical analysis on the combined dataset from all hospitals would in many cases yield more accurate results.

In this paper, we seek to circumvent these privacy issues that occur when multiple data owners engage in joint computations. More specifically, we aim to fit a Gaussian mixture model (GMM) based on a dataset that is distributed among a set of nodes. Hence, the nodes want to fit a GMM for the combined dataset (since using as much data as possible increases the accuracy of the model) without having to disclose their individual datasets. For training the model we consider the Expectation Maximization (EM) algorithm \cite{dempsterEM} because of its wide applicability.
Thus, there are a substantial amount of methods that are based on the EM algorithm (see for instance \cite{DistEM1, distEM2, distEM3}) where most would immediately be privacy-preserving by using a privacy-preserving distributed EM algorithm. 

The concept of \textit{federated learning} \cite{fedLearning} is widely used as a method to train a global model over multiple nodes without sharing each node's local data directly. It does so by each node training a local model and subsequently all local models are combined to a global model. In \cite{hitaj2017deep} it is shown how an adversary could attack this approach to learn private data, meaning that even though the nodes does not share directly their local data, information about this private data is leaked anyway. Therefore, we follow up on this result and use an information theoretic approach to prove that a federated version of the EM algorithm is indeed not privacy-preserving and subsequently we propose a solution to circumvent the situation.

There is a substantial amount of work dealing with privacy in machine learning methods, see for instance \cite{PrivMachine1, PrivMachine2, PrivMachine3, privMachine4, privMachine5, privMachine6,mcmahan2017communication,abadi2016deep}.
Of particular interest to our work is \cite{PrivEM1, privEM2, privEM3, privEM4, difEM1, difEM2, oblPoly} that also considers privacy of distributed EM. Among them, the privacy of the proposed solution in \cite{oblPoly} is based on two party cryptographic computations that tend to be computationally heavy and time consuming. The remaining approaches can be broadly classified into three types. (1) Homomorphic encryption based approaches: the proposed solution in \cite{privEM2, privEM3, privEM4} is based on homomorphic encryption which is known to be very computationally demanding and consequently time consuming. A distributed approach was proposed in \cite{privEM2} to reduce the overhead in computations. However, there is still a need for a lightweight solution.   (2) Differentially private approaches: the works \cite{difEM1, difEM2} propose differential private EM algorithms, however these approaches are not proposed in a distributed/decentralized setting. Additionally, there is an inevitable trade-off between privacy and accuracy of the output in differentially private approaches.
(3) Secure summation approach: the main idea of the work \cite{PrivEM1} is to apply a secure summation protocol in each M step for protecting privacy. 
The drawback is that it requires to first detect a so-called Hamiltonian cycle. In addition, this algorithm is very vulnerable to security attacks since it depends on a single node to keep the encryption seed.  

In order to address the limitations of the above-existing approaches,  in this paper, we propose a new privacy-preserving distributed EM algorithm for GMM using subspace perturbation \cite{Jane2020ICASSP,Jane2020TSP}, a  privacy-preserving technique based on distributed convex optimization.  By inspecting the updating functions of EM algorithm, we observe that the M step is the cause of privacy concern as it requires information exchange between different nodes.  To address it, we first formulate the required updates in M step as a distributed convex optimization problem and then exploit the subspace perturbation technique to securely compute these updates. We conduct numerical experiments to validate that the proposed approach achieves a higher privacy level compared to existing approaches without compromising the estimation accuracy.

\section{Preliminaries and problem setup}
The following notation is used in this paper. Lowercase letters $(x)$ denote scalars, lowercase boldface letters $(\v x)$ vectors,  uppercase boldface letters $(\v X)$ matrices, and calligraphic letters $(\mathcal{X})
$ sets. $\v x_{i}$ denotes the $i$-th entry of the vector $\v x$, and $\v X_{ij}$ denotes the $(i,j)$-th entry of the matrix $\v X$.  We use  an uppercase overlined letter $(\bar{X})$ to denote a random variable. With a slight abuse of notation we use $(\bar{X})$ for all cases when the realization is a scalar $(x)$, vector $(\v x)$, or matrix $(\v X)$ (can be distinguished from the context). 

\subsection{Fundamentals of GMM and EM over networks}
A network can be modelled as a graph (undirected) $\mathcal{G}=\{\mathcal{V},\mathcal{E}\}$ where $\mathcal{V}=\{1,\ldots,n\}$ denotes the set of $n$ nodes and $\mathcal{E} \subseteq \mathcal{V} \times \mathcal{V}$ denotes the set of $m$ edges. Note that node $i$ and $j$ can communicate with each other only if there is an edge connecting them, i.e., $(i,j)\in \mathcal{E}$. Denote $\mathcal{V}_i=\{j|(i,j)\in \mathcal{E}\}$ as the set of neighboring nodes of node $i$. We now introduce fundamentals about the GMM. Consider the scenario where $n$ nodes each has its own data $\v {x}_{i}$ and these nodes would like to collaborate to learn a GMM based on the full dataset $\{\v x_1,\v x_2, \ldots, \v x_n\}$. 
We start by introducing the GMM and explain how to fit it to the given dataset. 
Assume there are in total $c$ Gaussian models and denote $\mathcal{C}=\{1,\ldots,c\}$. Specifically, the GMM density is given by
\begin{equation}
    p(\v x) = \sum_{j\in \mathcal{C} } \beta_{j} p(\v x | \v \mu_j, \v \Sigma_j), 
\end{equation} 
where $p(\v x | \v \mu_j, \v \Sigma_j)$ is the pdf for a Gaussian distribution with mean $\v \mu_j$ and covariance matrix $\v \Sigma_j$ and the $\beta_{j}$'s are so-called mixture coefficients.  The task is then to estimate $\beta_{j}$, $\v  \mu_j$ and $\v \Sigma_{j}$ for all $j\in \mathcal{C}$, which can be done using the EM algorithm and the available data. The EM algorithm is iterative and for each iteration $t\in \mathcal{T}$ where $\mathcal{T}=\{0,1,\ldots,T\}$, the following steps are taken for all $j\in \mathcal{C}$:

\textbf{E-step:}
\begin{align}\label{eq.Estep}
    P(\v x_i | \N^{t}_j) = \frac{p(\v x_i | \v \mu_j, \v \Sigma_j) \beta_j^t }{\sum_{k=1}^c p(\v x_i | \v \mu_k, \v \Sigma_k) \beta_k^t },
\end{align}

\textbf{M-step:}
\begin{equation} \label{eq.Mstep}
    \begin{aligned}
    \beta_{j}^{t+1} &= \frac{\sum_{i=1}^n P(\v x_i | \N^{t}_j)    }{n} \\
    \v \mu_j^{t+1} &= \frac{\sum_{i=1}^n P(\v x_i | \N^{t}_j) \v x_i   }{\sum_{i=1}^n P(\v x_i | \N^{t}_j) } \\
    \v \Sigma_{j}^{t+1} &= \frac{\sum_{i=1}^n P(\v x_i | \N^{t}_j) (\v x_i - \v \mu_j^t)(\v x_i - \v \mu_j^t)^{\top}   }{\sum_{i=1}^n P(\v x_i | \N^{t}_j) }, 
    \end{aligned}
\end{equation}
where $ P(\v x_i | \N^{t}_j)$ denotes the conditional probability that data $\v x_i$ belongs to Gaussian model $j$.
In order to compute the updates in \eqref{eq.Mstep} in a distributed manner, data aggregation over the network is required. One straightforward way is for each node to publish its individual dataset to all neighboring nodes. However, directly publishing/sharing the individual data would violate the privacy.   

\subsection{Definition of private data and adversary models}
We define the private data to be the individual data held by each node as it often contains sensitive information. For example, a person's health condition, like whether he/she has Parkinson's disease or not, can  be revealed by voice data \cite{alavijeh2019quality}. 
Hence, each node $i$ would like to prevent that information regarding its own data $\v {x}_{i}$ is revealed to other nodes during the computation.

We also need to define what the private data should be protected against. We consider two well-known adversary models here: the eavesdropping adversary and  the passive (also called honest-but-curious) adversary. The eavesdropping adversary works by listening to the messages transmitted along the communication channels. The passive adversary model controls a number of so-called corrupt nodes who are assumed to follow the algorithm instructions but can collect information together. It will then use the collected information to infer the private data of the non-corrupt nodes, which we will refer to as honest nodes.

\subsection{Privacy-preserving distributed EM algorithm}\label{subsec.reqM}
Putting things together,  a privacy-preserving distributed EM algorithm for GMM should satisfy the following two requirements:
\begin{itemize}
 \item Individual privacy: throughout the algorithm execution, each honest node's private data should be protected from being revealed to the adversaries. We quantify the individual privacy using mutual information $I(\bar{X};\bar{Y})$,  which measures the mutual dependence between two random variables $\bar{X},\bar{Y}$. We have that $0\leq I(\bar{X};\bar{Y})$, with equality if and only if $\bar{X}$ is independent of $\bar{Y}$. On the other hand, if $\bar{Y}$ uniquely determines $\bar{X}$ we have $I(\bar{X};\bar{Y})=I(\bar{X};\bar{X})$ which is maximal.
    \item Output correctness:  the fitted model, i.e., the estimated parameters of GMM,  should be the same as  using non-privacy-preserving counterparts. Namely, the performance of the GMM  should not be compromised by considering privacy.
\end{itemize}

\section{Federated EM and its privacy leakage}
In what follows, we will use EM algorithm for GMM as an example to show that federated learning does not always guarantee privacy. 
\subsection{Federated EM algorithm for GMM}
Federated learning aims to learn the model under the constraint that the data is stored and processed locally, with only intermediate updates being communicated periodically with a central server. Therefore, in the context of EM algorithm, instead of directly sharing the data $\v {x}_{i}$, each node $i$ shares the following intermediate updates only:
\begin{equation}\label{eq.interM}
\begin{aligned}
     a_{ij}^t&=P(\v x_{i}|\N^{t}_j)\\ 
     \v b_{ij}^t&=P(\v x_{i}|\N^{t}_j)\v x_{i}\\
    \v C_{ij}^t&= P(\v x_{i}|\N^{t}_j)(\v x_{i}- \v \mu_j^t)(\v x_{i}-\v \mu_j^t)^\top
\end{aligned}
\end{equation}
Hence, all the above updates can also be computed locally at node $i$.
After receiving these intermediate updates from all nodes, the server will then aggregate all local updates and compute the global update $\beta_{j}^{t+1},\v \mu_j^{t+1}, \Sigma_{j}^{t+1}$ required in the M-step through the following way:
\begin{equation}\label{eq.update_sum}
\begin{aligned}
    \beta_{j}^{t+1}&=\frac{\sum_{i=1}^n a_{ij}^t}{n}=\frac{ a_{j}^t}{n}\\
    \v \mu_j^{t+1}&=\frac{\sum_{i=1}^n \v b_{ij}^t}{\sum_{i=1}^n a_{ij}^t}=\frac{\v b_{j}^t}{a_{j}^t}\\
    \v \Sigma_{j}^{t+1}&=\frac{\sum_{i=1}^n \v C_{ij}^t}{\sum_{i=1}^n  a_{ij}^t}=\frac{\v C_{j}^t}{ a_{j}^t}.
\end{aligned}
\end{equation}
Thereafter the server sends the global update $\beta_{j}^{t+1}, \v \mu_j^{t+1}, \v \Sigma_{j}^{t+1}$  back to each and every node. 

\subsection{Privacy leakage in Federated EM algorithm}\label{subsec:LeakFL}
We note that with the above federate EM algorithm, even though each node does not share the private date directly, it is still revealed to the server. This is because with the intermediate updates $a_{ij}^t$ and $\v b_{ij}^t$, the server is able to determine the private data $\v {x}_{i}$ of each node $i$ since $\v b_{ij}^t = a_{ij}^t\v {x}_{i}$. That is,  at each iteration the server has the following  mutual information 
\begin{align}\label{eq.fEM_exemplified}
   I(\bar{X}_{i};\bar{A}_{ij}^t, \bar{B}_{ij}^t )=I(\bar{X}_i,\bar{X}_i),
\end{align}
which is maximal. This means that every node's private data $\v x_{i}$ is completely revealed to the server. Hence, in this context federated EM algorithm is not privacy-preserving at all.

\section{Proposed approach} 
We now proceed to introduce the proposed approach which addresses the privacy issue raised in federated EM algorithm. As shown in the previous section, the exchange of intermediate updates $a_{ij}^t, \v b_{ij}^t, \v C_{ij}^t$ will reveal all private date of each node. 
We observe that it is sufficient to use the average updates $\frac{1}{n} a_{j}^t$, $\frac{1}{n}\v b_{j}^t$, and $\frac{1}{n}\v C_{j}^t$, to compute all global updates $\beta_j^{t+1}, \v \mu_j^{t+1}, \v \Sigma_j^{t+1}$ in \eqref{eq.update_sum}. Therefore, we propose to apply the subspace perturbation technique \cite{Jane2020ICASSP,Jane2020TSP} to securely compute these average updates without revealing each node's intermediate updates. In what follows we will first introduce fundamentals of subspace perturbation and then explain details of the proposed approach.

\subsection{Problem formulation using distributed convex optimization}
Assume $n$ nodes each has private data ${\v s}_i$ and  let $\v y_i$ be the so-called optimization variable, and stacking them together we have $\bs, \by\in \mathbb{R}^{n}$. The average consensus can  be formulated as a distributed convex optimization problem given by
\begin{equation} \label{eq.setupAve}
\begin{array}{ll}{\displaystyle \min_{\v y_i}} & {{\displaystyle\sum_{i \in \mathcal{V}} \frac{1}{2}\|\v y_i - {\v s}_i\|^2_2}} \\ {\text { s.t. }} & \quad {\v y_i=\v y_j}, {\forall} (i,j)\in \mathcal{E}, \end{array}
\end{equation}
where the optimum solution is
${\forall i\in \mathcal{V}: \v y_i^{*}=n^{-1} \sum_{i\in \mathcal{V}} {\v s}_i}$. With PDMM \cite{zhang2018distributed}, for each $e_l=(i,j)\in\mathcal{E}$ it defines two dual variables: ${{\v \lambda}_{l}={\v \lambda}_{i|j}}$, ${{\v \lambda}_{l+m}={\v \lambda}_{j|i}}$. 
The local updating functions are given by
\begin{align}
&\v y_i^{(t+1)} = \frac{\v s_i + \sum_{j\in N_i} \left(c\v y_j^{(t)} - \v B_{i|j}\v \lambda^{(t)}_{j|i} \right)}{1+cd_i} , \label{eq:xup}\\
&\forall j \in \mathcal{V}_i:  \v \lambda_{i|j}^{(t+1)} = \lambda_{j|i}^{(t)} + c\v B_{i|j} \left(\v y_i^{(t+1)} - \v y_j^{(t)}\right), \label{eq:lup}
\end{align}
where $c$ is a constant for controlling the convergence rate, $d_i=|\mathcal{V}_i|$ is the number of neighboring nodes of node $i$. The matrix $\bB \in \mathbb{R}^{m \times n}$ is related to the graph incidence matrix: $\bB_{li}=\bB_{i|j}$, $\bB_{lj}=\bB_{j|i}$ where $\bB_{i|j}=1$ if $i>j$ and $\bB_{i|j}=-1$ if  $i<j$.  
\subsection{Subspace perturbation and the proposed approach}
It is shown that the dual variable ${\v \lambda}^{(t)}$ will only converge in a subspace, denoted as $H$, determined by the graph incidence matrix. Denote $\Pi_{H}$ as the orthogonal projection into $H$.
We have ${\v \lambda}^{(t)}= (\Pi_{H})\v\lambda^{(t)}+ (\v I-\Pi_{H})\v\lambda^{(t)}$  where the former will converge to a fixed point while the latter will not.
The main idea of subspace perturbation is to exploit the non-convergent component of dual variable, i.e., $(\v I-\Pi_{H)}\v\lambda^{(t)}$ as subspace noise for guaranteeing the privacy. This can be achieved by initializing $\v\lambda^{(0)}$ with sufficiently large variance for protecting  the  private data  $\v s_i$ from being revealed to others. Additionally, the accuracy is guaranteed as the subspace noise has no effects on the convergence of the optimization variable.
Details of privacy-preserving distributed average consensus using subspace perturbation is summarized in Algorithm \ref{alg.1}. As mentioned before, the main idea of the proposed approach is to use subspace perturbation to securely compute the average results $\frac{1}{n} a_{j}^t$, $\frac{1}{n}\v b_{j}^t$, and $\frac{1}{n}\v C_{j}^t$ at each M step. Overall, we summarize details of the proposed approach in Algorithm \ref{alg.2}.

\subsubsection{Performance analysis}
Since the applied subspace perturbation output accurate average result as the non-privacy-preserving counterparts, it follows that the proposed approach also satisfies the output correctness requirement as the updates computed in M step is kept accurate. 

As for the individual privacy requirement, we will discuss it based on the adversary types. When dealing with the passive adversary, the subspace perturbation approach guarantees the privacy of each honest node as long as it has one honest neighboring node, i.e., $\mathcal{V}_i\cap\mathcal{V}_h\neq \emptyset$ where $\mathcal{V}_h$ denotes the set of honest nodes. 
In this situation, it is shown that the subspace perturbation protocol only reveals the sum of the honest nodes' inputs, assuming the honest nodes are connected after removing all corrupted nodes (see Proposition 3 in \cite{Jane2020TIFS}). That is, for each honest node $i\in \mathcal{V}_h$ its individual privacy at each iteration $t$ is given by
\begin{equation}\label{eq.indPro}
    I(\bar{X}_{i}; \{ \sum_{j\in \mathcal{V}_h}\bar{A}_{jk}^t, \sum_{j\in \mathcal{V}_h}\bar{B}_{jk}^t,  \sum_{j\in \mathcal{V}_h}\bar{C}_{jk}^t\}_{k \in \mathcal{C}}),
\end{equation}
which is a significant improvement compared to the \eqref{eq.fEM_exemplified} of the federated EM approach. As for the eavesdropping adversary, it requires no channel encryption except for the initialization step for transmitting $\v \lambda^{(0)}$.

\begin{algorithm}[t]
\vskip -2pt
\caption{Privacy-preserving distributed average consensus using subspace perturbation \cite{Jane2020TSP}}
\vspace{1.2mm}
\begin{algorithmic}[1] \label{alg.1}
\STATE Every node $i \in \mathcal{V}$ initializes $\v y_i^{(0)}$ arbitrarily and $\{{\v \lambda}_{j|i}^{(0)}\}_{j \in \mathcal{V}_{i}}$ based on the desired privacy level.
\STATE Node $i$ sends ${\v \lambda}_{i|j}^{(0)}$ to its neighbor $j \in \mathcal{V}_{i}$ through securely encrypted channels \cite{dolev1993perfectly}.
\WHILE {$\|\v y^{(t)} - \v y^*\|_2 < $ threshold}
\STATE Randomly activate a node with uniform probability, say node $i$, updates  ${\v y}_{i}^{(t+1)}$ using \eqref{eq:xup}.
\STATE Node $i$ broadcasts ${\v y}_{i}^{(t+1)}$  to its neighbors $j\in \mathcal{V}_i$.
\STATE All neighboring nodes $j \in \mathcal{V}_i$ update ${\v \lambda}_{j|i}^{(t+1)} $ using \eqref{eq:lup}.
\ENDWHILE
\end{algorithmic}
\vskip -2pt
\end{algorithm}

\begin{algorithm}[t]
\vskip -2pt
\caption{Proposed privacy-preserving distributed EM algorithm for GMM using subspace perturbation}
\vspace{1.2mm}
\begin{algorithmic}[1] \label{alg.2}
\STATE Initialize $\{ \beta_{j}^{0}, \v \mu_j^{0},\v \Sigma_{j}^{0}\}_{j\in \mathcal{C}}$
\WHILE {iteration $t\in {0,1,\ldots,T}$}
\STATE Each node $i$ first computes \eqref{eq.Estep} and then updates $a_{ij}^t,\v b_{ij}^t,\v C_{ij}^t$ using \eqref{eq.interM}.
\STATE Apply Algorithm \ref{alg.1} to securely compute the average results $\frac{1}{n} a_{j}^t$, $\frac{1}{n}\v b_{j}^t$ and $\frac{1}{n}\v C_{j}^t$ over the whole network. 
\STATE Each node $i$ updates $ \beta_j^{t+1}, \v \mu_j^{t+1}, \v \Sigma_j^{t+1}$ using \eqref{eq.update_sum}.
\ENDWHILE
\end{algorithmic}
\vskip -2pt
\end{algorithm}

\subsubsection{Comparison with the existing approach}\label{sec.com}
We remark that, among the existing approaches introduced in the introduction, the secure summation approach \cite{PrivEM1} is particularly comparable to our proposed solution. Because it is also a fully decentralized protocol that aims to achieve privacy without compromising the accuracy of the output (unlike the differentially private approaches). In addition, the privacy-preserving tool is not based on computationally complex functions such as homomorphic encryption. Its main idea  is to apply a secure summation protocol at each M step to first compute the sums, i.e.,  $\forall j \in \mathcal{C}:~ \sum_{i=1}^{n} a_{ij}^t, \sum_{i=1}^{n}\v b_{ij}^t, \sum_{i=1}^{n}\v C_{ij}^t$ and then updates $\beta_{j}^{t+1}, \v \mu_j^{t+1}, \v \Sigma_{j}^{t+1}$ using \eqref{eq.update_sum}. However, we remark the proposed approach achieves a higher privacy level than \cite{PrivEM1} when considering the passive adversary.  

To exemplify this claim, we consider the graph in Fig. \ref{fig_cycle}.
\begin{figure}
    \centering
    \begin{tikzpicture}[scale=0.50]
    \node[shape=circle,draw=black] (A) at (0,0) {$1$};
    \node[shape=circle,draw=red] (B) at (2,-1.5) {$2$};
    \node[shape=circle,draw=black] (C) at (4,0) {$3$};
    \node[shape=circle,draw=red] (D) at (3,2) {$4$};
    \node[shape=circle,draw=black] (E) at (1,2) {$5$};

    \path[blue] [-] (A) edge node[left, red] {} (B);
    \path [-] (A) edge node[left] {} (C);
    \path[blue] [-](B) edge node[left] {} (C);
    \path [-](B) edge node[left] {} (D);
    \path[blue] [-](D) edge node[left] {} (C);
    \path[blue] [-](A) edge node[right] {} (E);
    \path[blue] [-](D) edge node[left] {} (E);
\end{tikzpicture}
    \caption{Example with cycle. The blue edges form a Hamiltonian cycle. The red nodes are corrupt nodes.}
    \label{fig_cycle}
\end{figure}
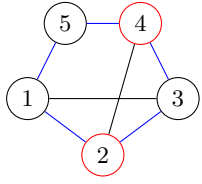
The secure summation protocol works by first detecting a  Hamiltonian cycle in the graph, which is a path in the graph visiting all the nodes once, and where the start and end node is the same.
Take the process of computing the sum $\sum_{i=1}^{n}a_{ij}^t$ as an example, node $1$ chooses a random number $r$ and then sends $a_{1j}^t+r$ to node $2$. Node $2$ adds $a_{2j}^t$ and send this to node $3$, repeat until node $5$ sends $\sum_{i=1}^{5}a_{ij}^t+r$ back to node $1$ which subtracts $r$ and broadcast $\sum_{i=1}^{5}a_{ij}^t$ to all nodes. 
Since the passive adversary is able to collect information from corrupted node $2$ and $4$, it has the following information denoted as 
\begin{align*}
   \mathcal{V}^{t}_{aj}&=\{\bar{A}_{1j}^t+\bar{R},\bar{A}_{2j}^t,\bar{A}_{1j}^t+\bar{A}_{2j}^t+\bar{A}_{3j}^t+\bar{R},\bar{A}_{4j}^t, \sum_{i=1}^{5}A_{ij}^t\}.
\end{align*}
Remark that $\bar{A}_{3j}^t=(\bar{A}_{1j}^t+\bar{A}_{2j}^t+\bar{A}_{3j}^t+\bar{R})-(\bar{A}_{1j}^t+\bar{R})-\bar{A}_{2j}^t$  Hence $\bar{A}_{3j}^t$ can be uniquely determined using the information in $\mathcal{V}^t_{aj}$.

Similarly, the adversary can collect information $\mathcal{V}^{t}_{bj}$ and $\mathcal{V}^{t}_{cj}$ when computing $\sum_{i=1}^{n}\v b_{ij}^t$ and $ \sum_{i=1}^{n}\v C_{ij}^t$, respectively. Overall, we conclude that at each iteration all information collected by the passive adversary is given by $\mathcal{V}^t=\{\mathcal{V}^{t}_{aj},\mathcal{V}^{t}_{bj},\mathcal{V}^{t}_{cj}\}_{j \in \mathcal{C}}$.
Therefore, for honest node $3$, its individual privacy is given by 
\begin{equation}
   I(\bar{X}_{3}; \{ \bar{A}_{3k}^t, \bar{B}_{3k}^t,  \bar{C}_{1k}^t\}_{k \in \mathcal{C}})=I(\bar{X}_{3};\bar{X}_{3})\label{eq:inf_leak2} \\
\end{equation}
As for honest nodes $i=1,5$, we have 
\begin{align}
    I(\bar{X}_i;\mathcal{V}^t)&\geq I(\bar{X}_{i}; \{ \sum_{j=1,5}\bar{A}_{jk}^t, \sum_{j=1,5}\bar{B}_{jk}^t,  \sum_{j=1,5}\bar{C}_{jk}^t\}_{k \in \mathcal{C}})\label{eq:inf_leak1}. 
\end{align}
As for the proposed approach, inserts $\mathcal{V}_h=\{1,3,5\}$ in  \eqref{eq.indPro} would yield the individual privacy of honest node $1,3,5$,
which is less than \eqref{eq:inf_leak2}, \eqref{eq:inf_leak1}, thereby proving our claim that the proposed approach achieves a higher privacy level than the existing approach \cite{PrivEM1}.
\section{Numerical results}
In this section, we demonstrate numerical results to validate the comparisons shown in the above section.
\subsection{Output correctness} \label{subsec.numout} 
We first simulated a geometrical graph with $n=80$ by allowing every two nodes to communicate if and only if their distance is within a radius of $\sqrt{2\log n/n}$, thereby ensuring a connected graph at a high probability $1-{1}/{n^2}$ \cite{dall2002random}. 
We use the Parkinsons dataset \cite{little2007exploiting} from the UCI repository \cite{Dua:2019}. This dataset contains voice measurements from 31 people and 23 of them are with Parkinson's disease. There are 195 instances in total and each has 22 features. The reason for choosing this dataset is that the involved biomedical voice measurements are highly sensitive information. In the experiment, to reduce the dimensionality of the features we apply principle component analysis to the dataset and choose the first two principle components for GMM fitting. 
As shown in Fig. \ref{fig.out}, we see that the proposed approach estimates the parameters for GMMs identically to the existing approach \cite{PrivEM1} which has perfect output correctness. Hence, the output correctness of the proposed approach is guaranteed. 
\begin{figure}[t]
\centering
\includegraphics[width=.35\textwidth]{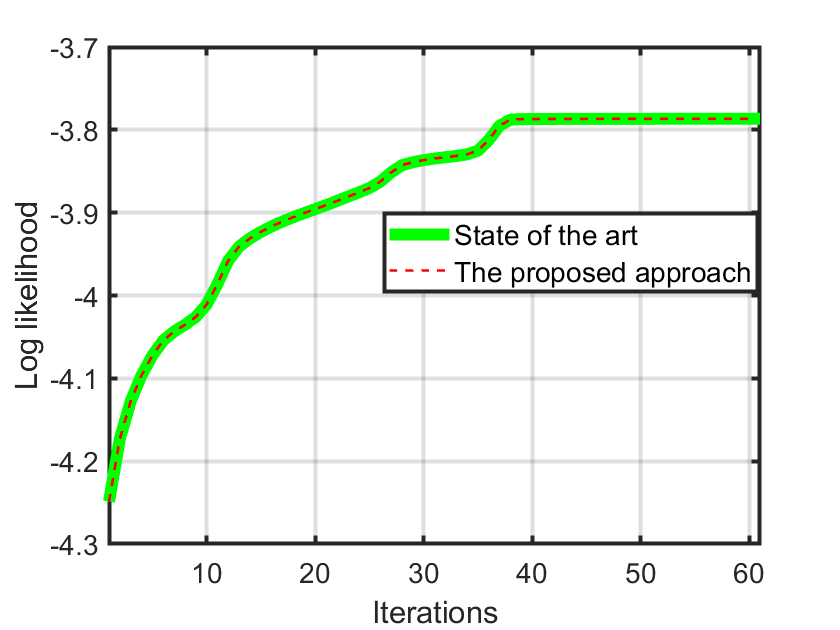}
 \centering
\caption{Log likelihood as a function of the iteration number of the proposed approach and the existing approach.}
\label{fig.out}
\vskip -10pt
\end{figure}
\subsection{Individual privacy}\label{subsec.numind}
Previously, we have proved that the federated EM algorithm is not privacy-preserving and the  privacy level of the proposed approach is higher than the existing approach \cite{PrivEM1} considering the passive adversary. To validate these results, we use the sample case shown in Fig. \ref{fig_cycle}. More specifically, we would like to show the individual privacy of honest node $1$ given corruptions in the neighborhood using these three algorithms. In order to demonstrate an accurate mutual information estimation, it requires to gather a massive amount of data for each node as statistical analysis needs to be performed. As a consequence, the above deployed Parkinson's dataset is too small to conduct mutual information estimation. To address it, we first generate synthetic data by assuming all the private data $\bar{X}_1, \ldots,\bar{X}_5$ are Gaussian distributed with zero mean and unit variance.  Additionally, all $\bar{A}_{ij}$'s are assumed uniformly distributed and their sum is normalized to one. 
After that, we perform $10^4$ times Monte Carlo simulations and then deploy npeet \cite{ver2000non} to estimate the normalized mutual information of honest node 1. The results are shown in Fig. \ref{fig.ind}. 
We can see that as expected, federated EM algorithm reveals all private information and among all approaches the proposed one reveals the minimum amount of information. 
\begin{figure}[t]
\centering
\includegraphics[width=.35\textwidth]{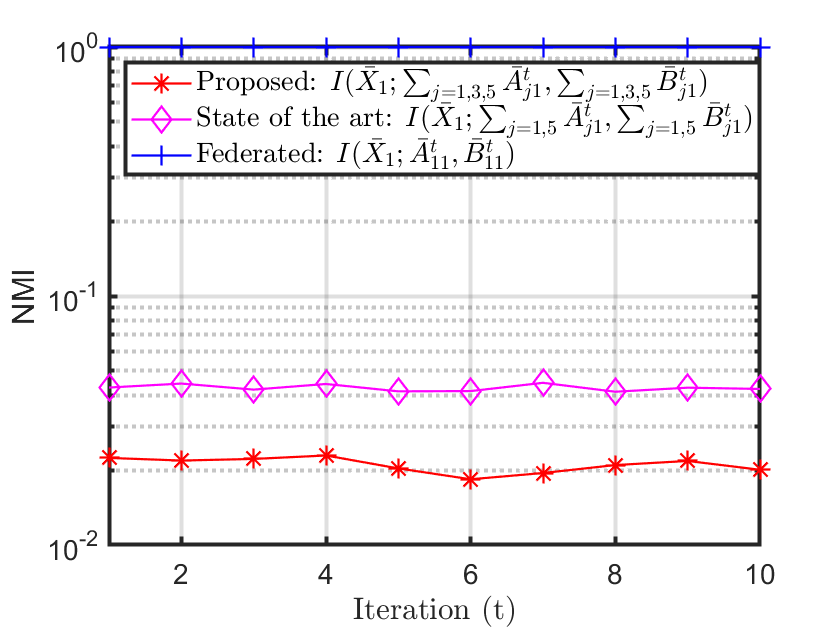}
 \centering
\caption{Normalized mutual information (NMI), i.e., individual privacy of honest node $1$ in terms of iteration number using the Federated, the existing and the proposed approach.}
\label{fig.ind}
\vskip -10pt
\end{figure}

\section{Conclusion}
In this paper, we consider the problem of privacy-preserving distributed EM for GMM.  We first gave explicit information-theoretical privacy analysis to prove that a federated EM algorithm does not guarantee privacy. After that, we proposed a lightweight privacy-preserving distributed EM algorithm for GMM which has no privacy-accuracy trade-off. Moreover, it offers stronger privacy guarantee when dealing with a number of corrupt nodes than the state-of-the-art algorithm. Numerical simulations were conducted to validate the above claims and demonstrate the superior performances of the proposed approach.

\newpage
\bibliographystyle{IEEEbib}
\bibliography{dualpath}
\end{sloppy}
\end{document}